\documentclass[conference]{IEEEtran}
\IEEEoverridecommandlockouts
\usepackage{url}
\usepackage{cite}
\usepackage{amsmath,amssymb,amsfonts}
\usepackage{amsthm}
\usepackage{algorithmic}
\usepackage{graphicx}
\usepackage{textcomp}
\usepackage{xcolor}
\usepackage{tabularx}   
\usepackage{booktabs}   
\usepackage{array}      
\usepackage{multirow, multicol}
\usepackage{enumitem}

\theoremstyle{plain}

\theoremstyle{definition}

\theoremstyle{remark}

\def\BibTeX{{\rm B\kern-.05em{\sc i\kern-.025em b}\kern-.08em
    T\kern-.1667em\lower.7ex\hbox{E}\kern-.125emX}}
\begin{document}

\title{HCOE: Hyperbolic Clinical Ontology Embeddings from Biomedical Language Models\\
\thanks{This work was supported in part by the NVIDIA Academic Grant Program and the Google Cloud Research Credits program.}
}

\author{
\IEEEauthorblockN{Yixuan Li}
\IEEEauthorblockA{\textit{McGill University}}
\and
\IEEEauthorblockN{Weihao Li}
\IEEEauthorblockA{\textit{Northwestern University}}
\and
\IEEEauthorblockN{Ziyang Song\textsuperscript{*}}
\IEEEauthorblockA{\textit{Ohio University}}
}



\maketitle

\begingroup
\renewcommand{\thefootnote}{}
\footnotetext{\textsuperscript{*}Corresponding author: \texttt{ziyangs@ohio.edu}. 
Accepted at the 2026 IEEE International Conference on Bioinformatics and Biomedicine (BIBM).}
\endgroup

\begin{abstract}

Biomedical language models (LMs) encode textual semantics but do not explicitly preserve medical code hierarchies. We present \textbf{Hyperbolic Clinical Ontology Embeddings (HCOE)} for hierarchy-aware clinical concept representation. HCOE maps frozen BioBERT embeddings into a Poincar´e ball, combining parent-side
and child-side ontology-guided contrastive learning with
coarse-to-fine ontology-path aggregation. It uses International Classification of Diseases (ICD) codes organized by Clinical Classifications Software (CCS) and Anatomical Therapeutic Chemical (ATC) medication hierarchies. Evaluations show that HCOE performs best on ICD/ATC clinical relation prediction and CCS-to-PheCode hierarchy transfer. On the MIMIC-IV dataset, HCOE also achieves the best  performance on mortality prediction, readmission prediction, medication recommendation, and rare drug prediction.
\end{abstract}

\begin{IEEEkeywords}
Clinical ontology embeddings, biomedical language models, hyperbolic representation learning, ontology-guided contrastive learning, electronic health records.
\end{IEEEkeywords}

\section{Introduction}
In electronic health record (EHR) modeling, diagnosis, procedure, and medication codes are used to represent patient visits and longitudinal histories for tasks such as diagnosis prediction, drug recommendation, and patient risk stratification \cite{TrajGPT, wang2026saferxagent, Yang2026-ls}. Biomedical language models (LMs) provide transferable textual representation of medical concepts \cite{lee2020biobert}, but their pretraining does not explicitly capture structural relationships among clinical concepts. The International Classification of Diseases (ICD) and Anatomical Therapeutic Chemical (ATC) classification organize diagnoses, procedures, and medications into multi-level clinical coding hierarchies, ranging from broad categories to fine-grained codes~\cite{icd, atc}. These clinical ontologies contain clinically meaningful coarse-to-fine relationships that are important for clinical concept representation \cite{shen2025smi}.

\begin{figure}[t]
    \centering
    \includegraphics[width=\columnwidth]{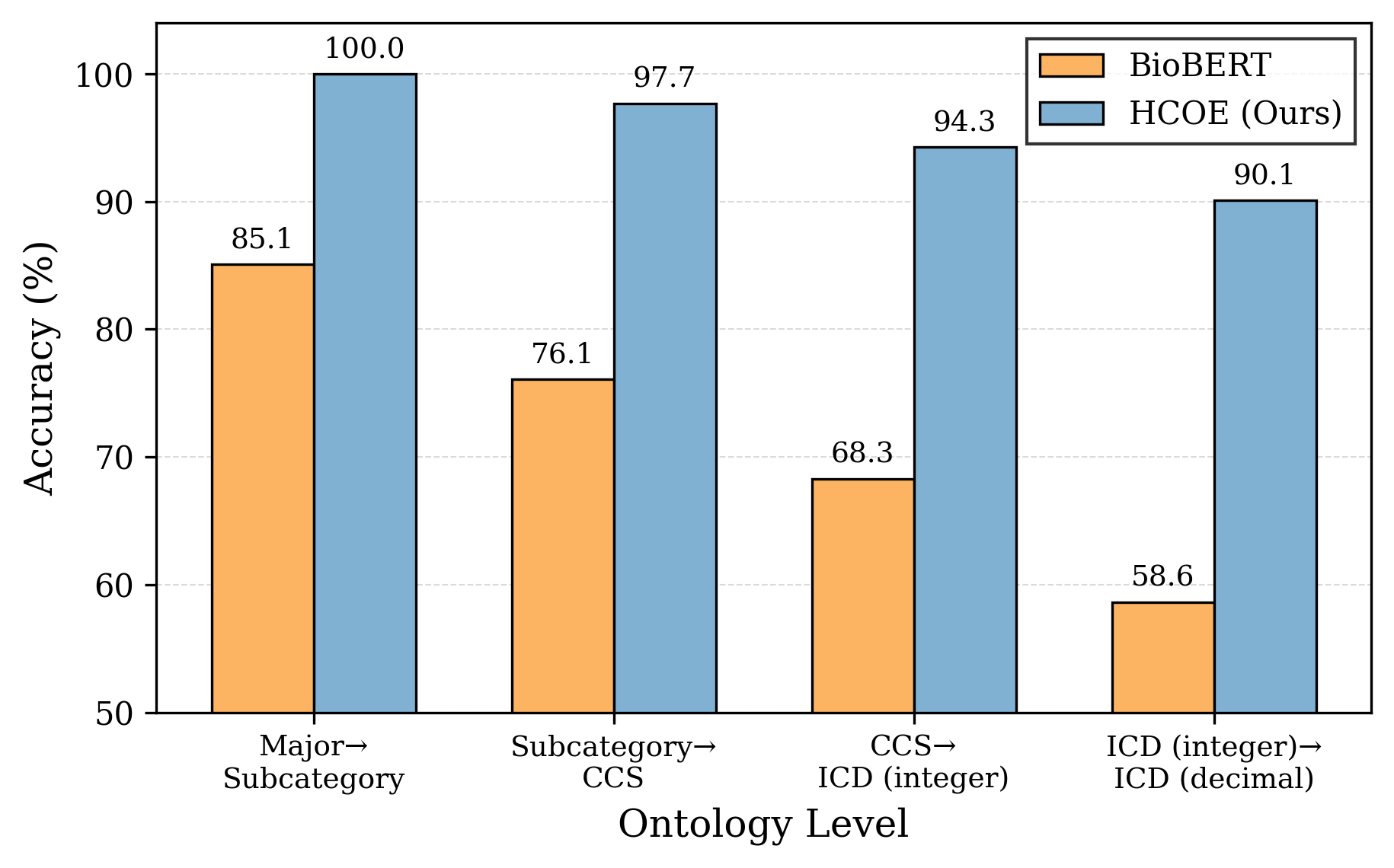}
    \caption{
    \textbf{Comparison of 1-hop parent–child relation prediction in the ICD  hierarchy.} We assess whether BioBERT and HCOE embed each child concept closer to its true parent than to unrelated concepts across five levels. HCOE consistently outperforms BioBERT, with larger gains at finer-grained levels.
    }
    \label{fig:motivation}
    \vspace{-1\baselineskip} 
\end{figure}

To motivate the need for hierarchy-aware clinical concept embeddings, we evaluate BioBERT, a biomedical LM pre-trained on PubMed abstracts~\cite{lee2020biobert}, on 1-hop parent--child relation prediction in the ICD hierarchy. The task tests whether each child concept is embedded closer to its true parent than to unrelated concepts. As shown in Fig.~\ref{fig:motivation}, BioBERT performs reasonably well at coarse hierarchy levels, but its accuracy decreases as the hierarchy becomes more fine-grained, reaching 58.6\% at the finest level. In contrast, our proposed method maintains higher accuracy across levels. These results suggest that biomedical LM semantics alone are insufficient to preserve fine-grained ICD hierarchy relations, motivating hierarchy-aware clinical concept embeddings that incorporate clinical ontology structure.

In EHR modeling, deep learning methods learn clinical code embeddings from co-occurrence patterns and longitudinal patient histories to support clinical prediction tasks 
\cite{TrajGPT, beam2020clinical}. To incorporate domain knowledge, several methods further extend this paradigm with clinical ontologies  or graph structures for more informative patient and concept  representations \cite{choi2017gram, shang2019pre, Yang2026-ls}. However, these methods are often trained from scratch for specific prediction tasks and do not exploit the transferable knowledge learned from large-scale biomedical corpora. Biomedical LMs address this limitation by pretraining on large-scale biomedical data, providing transferable textual semantics for medical concepts, but they are not explicitly optimized to distinguish hierarchical relations among clinical codes \cite{lee2020biobert, alsentzer2019publicly}. 
Recent hierarchy-aware LMs incorporate structured knowledge to learn more hierarchy-aware representations, but their Euclidean embeddings are less suitable for representing clinical ontologies  \cite{liu2021learning}. In addition, hyperbolic embedding methods provide a natural geometry for modeling hierarchical relations, but they generally do not leverage pre-trained LM representations \cite{nickel2017poincare, pmlr-v80-ganea18a, tifrea2018poincare}. This motivates integrating pre-trained biomedical LM representations with clinical ontologies in hyperbolic space for hierarchy-aware clinical concept representation.

We present \textbf{Hyperbolic Clinical Ontology Embeddings (HCOE)}, a hyperbolic representation learning framework for hierarchy-aware clinical concept embeddings. HCOE learns parent–child relations from CCS-organized ICD codes and ATC hierarchies using ontology-guided contrastive learning, then aggregates embeddings along each coarse-to-fine ontology path. We evaluate
clinical relation prediction, CCS-to-PheCode hierarchy transfer,
and downstream MIMIC-IV tasks. The main contributions
are:
\begin{enumerate}
    \item We develop HCOE for hierarchy-aware medical code embeddings from frozen biomedical LM representations and clinical ontologies.
    \item We introduce ontology-guided hyperbolic contrastive learning and coarse-to-fine Möbius path aggregation.
    \item We demonstrate the advantage of HCOE on ICD/ATC relation prediction, CCS-to-PheCode hierarchy transfer, and MIMIC-IV mortality prediction, readmission prediction, medication recommendation, and rare drug prediction.
\end{enumerate}

\section{Related Work}

Clinical concept and medical code representation has been widely studied for EHR modeling. Early methods learn medical code embeddings from co-occurrence patterns and longitudinal patient records, such as cui2vec~\cite{beam2020clinical}. To incorporate domain knowledge, ontology- and graph-enhanced representation methods further use clinical hierarchies or graph structures to improve clinical concept embeddings. For example, GRAM uses ontology ancestors with attention to learn clinically meaningful medical code representations~\cite{choi2017gram}, while RotatE represents relations in structured knowledge graphs as rotations in a complex embedding space and captures relational patterns among concepts~\cite{sun2019rotate}. However, these methods are trained from scratch on task-specific data and do not fully exploit transferable knowledge learned from large-scale corpora.

Biomedical LMs, such as BioBERT~\cite{lee2020biobert} and ClinicalBERT~\cite{alsentzer2019publicly}, provide transferable contextual semantics for medical concepts by pretraining on large-scale biomedical or clinical corpora and have shown strong performance on downstream clinical prediction tasks. However, these models primarily capture textual semantics and are not explicitly optimized to preserve hierarchical relations among clinical codes, limiting their ability to model clinical ontology structure. Recent hierarchy- and ontology-aware LM methods further use structured biomedical knowledge to improve clinical representations. For example, SapBERT uses UMLS-based synonym alignment to learn better clinical concept embeddings~\cite{liu2021learning}, while G-BERT combines EHR sequences with medical ontology graphs for medication recommendation~\cite{shang2019pre}. However, these methods learn embeddings in Euclidean space, which is less suitable for modeling the tree-structured hierarchies of clinical ontologies. In contrast, hyperbolic representation learning provides a more suitable geometry for hierarchical structures because the volume of hyperbolic space grows exponentially with radius. Methods such as Poincaré Embeddings~\cite{nickel2017poincare}, Hyperbolic Entailment Cones~\cite{pmlr-v80-ganea18a}, and Poincaré GloVe~\cite{tifrea2018poincare} have shown the effectiveness of hyperbolic spaces for modeling hierarchical or relational structures. However, these methods are not designed to integrate pre-trained biomedical LM semantics with expert-curated clinical ontologies. To bridge this gap, HCOE learns hierarchy-aware medical code embeddings by integrating biomedical LM semantics with clinical ontologies in hyperbolic space.

\section{Background of Hyperbolic Representation}

Hyperbolic representation learning embeds hierarchical structures in a negatively curved space, whose exponential volume growth naturally fits tree-like structures \cite{nickel2017poincare}. It motivates hierarchy-preserving hyperbolic embedding methods such as Poincaré embeddings \cite{nickel2017poincare}, Poincaré GloVe \cite{tifrea2018poincare}, and entailment cones \cite{pmlr-v80-ganea18a}. Here, we use the $d$-dimensional Poincaré ball with a curvature $-\kappa$ ($\kappa>0$):
\begin{equation}
    \mathbb{B}^d_\kappa = \left\{ \mathbf{x} \in \mathbb{R}^d : \kappa\,\|\mathbf{x}\|^2 < 1 \right\}
\end{equation}
The hyperbolic distance between two points $\mathbf{u}, \mathbf{v} \in \mathbb{B}^d_\kappa$ is:
\begin{equation}
    d_\kappa(\mathbf{u}, \mathbf{v})
    = \frac{2}{\sqrt{\kappa}} \tanh^{-1}\!\left(\sqrt{\kappa}\,\|(-\mathbf{u}) \oplus_\kappa \mathbf{v}\|\right)
\end{equation}
where $\|\cdot\|$ is Euclidean norm and $\oplus_\kappa$ is
M\"obius addition:
\begin{equation} \label{eq:mobius-add}
    \mathbf{u} \oplus_\kappa \mathbf{v} = \frac{(1 + 2\kappa\langle \mathbf{u}, \mathbf{v} \rangle + \kappa\|\mathbf{v}\|^2)\mathbf{u} + (1 - \kappa\|\mathbf{u}\|^2)\mathbf{v}}{1 + 2\kappa\langle \mathbf{u}, \mathbf{v} \rangle + \kappa^2\|\mathbf{u}\|^2\|\mathbf{v}\|^2}
\end{equation}
where $\langle \cdot, \cdot \rangle$ denotes the Euclidean inner product.

\section{Methodology}



\subsection{ICD and ATC Clinical Ontologies}
\label{sec: medical_onto}

Clinical coding systems, such as ICD and ATC, organize diagnoses, procedures, and medications into multi-level taxonomies from broad categories to fine-grained codes. For this study, we map each clinical concept to a unique five-level ontology path, which defines the parent-child relations used to train HCOE. 
For diagnoses, we use the Clinical Classifications Software (CCS) hierarchy developed by the Agency for Healthcare Research and Quality (AHRQ) to organize ICD-CM codes into five levels: major category, sub-category, CCS code, integer-level ICD-CM code, and decimal-level ICD-CM code. 
For procedures, we similarly organize ICD-PCS codes using the CCS procedure hierarchy into major category, sub-category, CCS code, three-character ICD-PCS prefix, and full ICD-PCS code. 
For medications, we use the five-level ATC hierarchy: Anatomical Main Group $\rightarrow$ Therapeutic Subgroup $\rightarrow$ Pharmacological Subgroup $\rightarrow$ Chemical Subgroup $\rightarrow$ Chemical Substance.

\begin{figure*}[t]
\centering
\includegraphics[width=\textwidth]{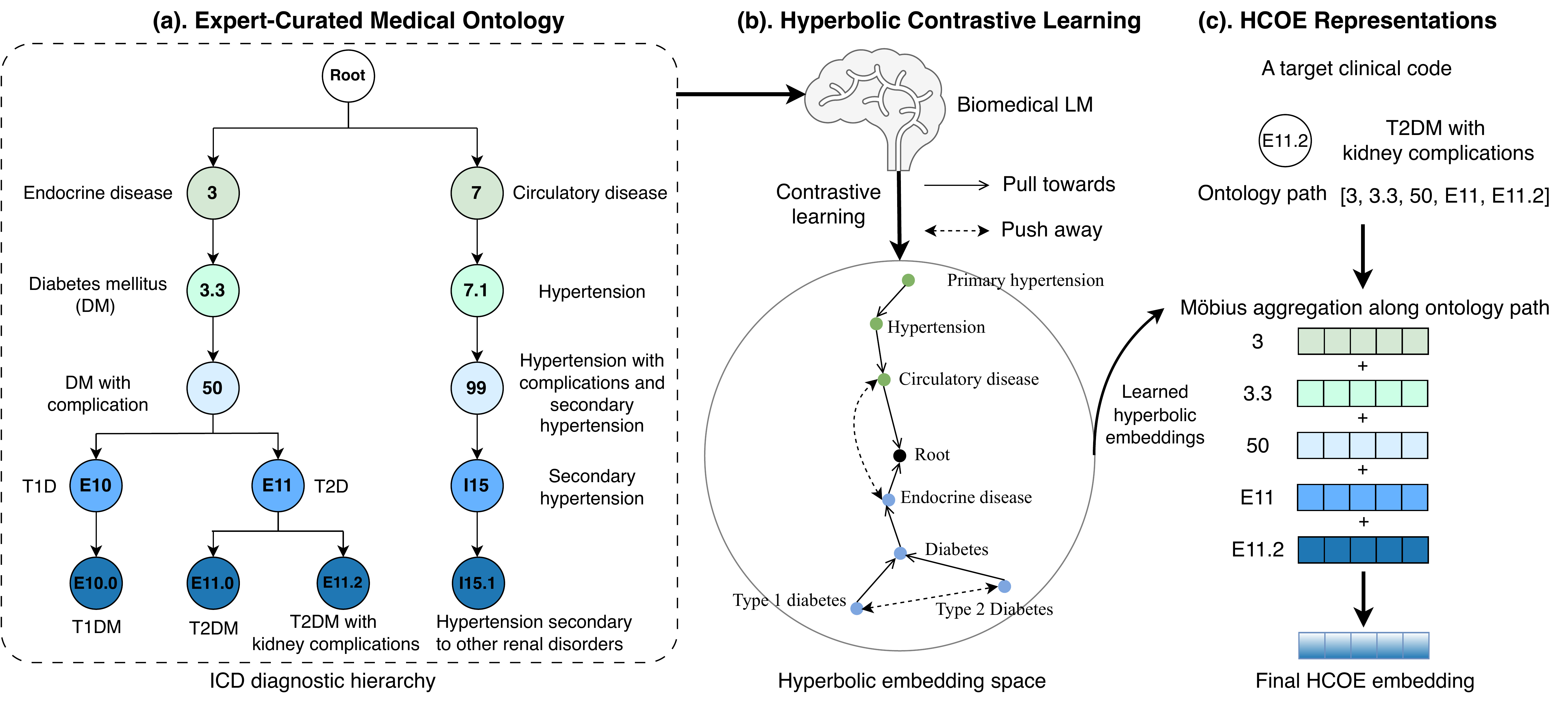} 
\caption{Overview of HCOE. \textbf{a}. An expert-curated medical ontology organizes clinical concepts into coarse-to-fine hierarchy levels. \textbf{b}. HCOE uses hyperbolic contrastive learning to organize biomedical LM-encoded clinical codes, pulling parent--child concepts closer and pushing unrelated concepts apart. \textbf{c}. For a target clinical concept, HCOE aggregates embeddings along its ontology path to form a coarse-to-fine hierarchy-aware representation. }
\label{fig: HCOE}
\vspace{-1\baselineskip} 
\end{figure*}

\subsection{Clinical Ontology Paths and LM Initialization}
\label{sec:ontology_path}

Clinical concepts are organized in an expert-curated medical ontology with $L$ levels of granularity. Let $C^{(\ell)}=\{c^{(\ell)}_{1},\dots,c^{(\ell)}_{|C^{(\ell)}|}\}$ denote the set of clinical concepts at level $\ell \in \{1,\dots,L\}$. For each clinical concept $i$, we define its ontology path as the ordered sequence of ancestor concepts from coarse to fine levels:
\begin{equation}
\mathcal{P}(i) \equiv \big(a_i^{(1)},\dots,a_i^{(L)}\big)
\end{equation}
where $a_i^{(\ell)} \in C^{(\ell)}$ denotes the ancestor of concept $i$ at level $\ell$. As shown in Fig.~\ref{fig: HCOE}c, the ICD-10-CM code \texttt{E11.2} (type 2 diabetes mellitus with kidney complications) follows the path \texttt{[3, 3.3, 50, E11, E11.2]} corresponding to endocrine disease (Major category), diabetes mellitus with complication (Sub-category), diabetes mellitus with complication (CCS code), type 2 diabetes mellitus (integer-level ICD), and itself (decimal-level ICD).

We construct a \textit{hierarchy-aware semantic embedding} for concept $i$ by aggregating level-specific embeddings of its ancestor codes based on its ontology path $\mathcal{P}(i)$. For each clinical code $c^{(\ell)}_j$ at level $\ell$, we use a frozen biomedical LM $\text{Enc}(\cdot)$ to encode its textual description and obtain its embedding from the pooled LM output:
\begin{equation}
\mathbf{x}^{(\ell)}_{j}
=
\mathrm{Pooler}\big(\mathrm{Enc}(c^{(\ell)}_{j})\big).
\end{equation}
Let $\mathbf{X}^{(\ell)} \in \mathbb{R}^{|C^{(\ell)}|\times d_0}$ denote the LM-initialized embedding table at level $\ell$, where $d_0$ is the output dimension of the biomedical LM. 
These embeddings provide semantic initialization for HCOE, while the ontology paths provide parent-child relations for hyperbolic representation learning.


\subsection{Hyperbolic Clinical Ontology Embeddings}
\label{sec: hcoe}

HCOE learns hyperbolic clinical ontology embeddings from biomedical LM representations using ontology-guided hyperbolic contrastive learning and coarse-to-fine path aggregation. Hyperbolic space is well suited to hierarchical clinical ontologies because it provides greater capacity for representing increasingly fine-grained concepts~\cite{nickel2017poincare}. The CCS/ICD and ATC hierarchies provide the parent-child relations and ontology paths used to train HCOE. We map frozen biomedical LM embeddings into a shared Poincaré ball using level-specific linear projections followed by the exponential map, with the projection parameters optimized through ontology-guided hyperbolic contrastive learning. For each ontology node $c_j^{(\ell)}$, we compute:
\begin{equation}
\mathbf{z}^{(\ell)}_j =
W_\ell \mathbf{x}^{(\ell)}_j + \mathbf{b}_\ell,
\qquad
\mathbf{h}^{(\ell)}_j = \exp^{\kappa}_{\mathbf{0}}\big(\mathbf{z}^{(\ell)}_j\big),
\end{equation}
where $W_\ell$ and $\mathbf{b}_\ell$ are level-specific projection parameters, $\exp^{\kappa}_{\mathbf{0}}(\cdot)$ is the exponential map at the origin, and $\mathbf{h}^{(\ell)}_j \in \mathbb{B}^{d_h}_{\kappa}$ is the hyperbolic embedding of ontology node $c^{(\ell)}_j$. As a result, all nodes are mapped to the same $d_h$-dimensional Poincaré ball, allowing parent-child distances to be computed directly.

To encode clinical hierarchy, we optimize HCOE using parent-child relations in the ontology. As shown in Fig.~\ref{fig: HCOE}b, we use a hyperbolic contrastive learning strategy that uses both parent-side and child-side triplet losses. The triplet losses encourage semantically related concepts to be close, while pushing unrelated ones farther apart. For a given anchor concept $u$, we construct a triplet $(u, p, n)$ by selecting a positive sample $p$ and a negative sample $n$. The resulting hyperbolic triplet loss is:
\begin{equation}
    \mathcal{L} = \sum_{(u, p, n)} \max \left( 0, \; d_\kappa(\mathbf{h}_u, \mathbf{h}_p) - d_\kappa(\mathbf{h}_u, \mathbf{h}_n) + \alpha \right)
\end{equation}
where $d_\kappa(\cdot,\cdot)$ denotes the hyperbolic distance in the Poincaré ball and $\alpha$ is a margin hyperparameter.

\begin{table*}[ht]
\centering
\setlength{\tabcolsep}{4.5pt}
\caption{Clinical relation prediction on ICD$\rightarrow$ICD, ATC$\rightarrow$ATC, ICD$\rightarrow$ATC, and CCS$\rightarrow$PheCode. Thresholds are selected on validation and fixed for testing without variance. The best results in each column are highlighted in bold.  }
\label{tab:hier_pred_prf}
\begin{tabular}{lccc ccc ccc ccc}
\toprule
\multirow{2}{*}{\textbf{Method}} &
\multicolumn{3}{c}{\textbf{ICD$\rightarrow$ICD}} &
\multicolumn{3}{c}{\textbf{ATC$\rightarrow$ATC}} &
\multicolumn{3}{c}{\textbf{ICD$\rightarrow$ATC}} &
\multicolumn{3}{c}{\textbf{CCS$\rightarrow$PheCode}} \\
\cmidrule(lr){2-4}\cmidrule(lr){5-7}\cmidrule(lr){8-10}\cmidrule(lr){11-13}
& \textbf{Precision} & \textbf{Recall} & \textbf{F-score}
& \textbf{Precision} & \textbf{Recall} & \textbf{F-score}
& \textbf{Precision} & \textbf{Recall} & \textbf{F-score}
& \textbf{Precision} & \textbf{Recall} & \textbf{F-score} \\
\midrule
\textbf{HCOE} & \textbf{76.3} & 74.9 & \textbf{75.6} & \textbf{85.1} & \textbf{69.3} & \textbf{76.3} & \textbf{54.2} & \textbf{59.4} & \textbf{56.7} & \textbf{85.4} & 76.3 & \textbf{80.6}\\
\midrule
BioBERT & 30.4 & \textbf{75.8} & 43.3 & 50.7 & 45.7 & 48.1 & 49.3 & 57.0 & 52.9 & 76.8 & 64.1 & 69.9\\
SapBERT & 69.9 & 64.5 & 67.1 & 84.0 & 66.9 & 74.5 & 50.2 & 55.7 &52.9
& 73.4 & 75.9 & 74.7\\
HiTs & 69.3 & 61.7 & 65.3 & 83.0 & 66.2 & 73.6 & 50.5 & 56.3 & 53.2 & 72.8 & \textbf{80.5} & 76.5  \\
OnT & 65.9 & 59.1 & 62.3 & 78.5 &62.3  & 69.5  &48.9 & 55.2 & 51.9 & 78.2 & 71.5 & 74.7\\
RotatE & 69.3 & 58.2 & 63.3 & 80.3 & 63.9 & 71.2 & 49.2 & 57.2 & 52.9 & 72.5 & 67.4 & 69.9 \\
cui2vec & 68.3 & 60.3 & 64.1 & 77.2 & 61.4 & 68.4 & 45.7 & 54.0 & 49.5  & 69.2 & 68.8 & 69.1 \\
PoincaréEmbed & 69.7 &  58.0 & 63.4 & 79.6 &60.7 & 68.9 & 43.2 & 52.1 & 47.2 & 65.1 & 79.2  & 71.5 \\
HyperbolicCone & 64.5 &69.2 & 66.8& 78.3 & 68.8 & 73.2 & 45.9 & 57.2 & 50.9 & 72.3 & 78.4 & 75.3 \\
PoincaréGloVe & 70.5 & 45.9 & 55.6 & 77.2 & 50.6 & 61.1 & 42.7 & 51.9 & 46.9 & 73.1 & 76.8 & 74.9 \\
\bottomrule
\end{tabular}
\end{table*}

We compute triplet losses from both the parent-side and the child-side. The parent-side objective uses a child concept as the anchor and its parent as the positive concept, encouraging each child to remain close to its ancestor. The child-side objective uses a parent concept as the anchor and one of its children as the positive concept, encouraging parent concepts to preserve their local descendant structure. Negatives are sampled from sibling concepts or unrelated concepts in the ontology. The final HCOE training objective is
\begin{equation}
    \mathcal{L}_{\text{HCOE}}
    = \mathcal{L}_{\text{parent}} + \mathcal{L}_{\text{child}}
\end{equation}

After training, each ontology node has a hyperbolic embedding $\mathbf{h}^{(\ell)}_j$. For a fine-grained clinical code $i$, we obtain its final representation by aggregating the hyperbolic  embeddings of nodes along its ontology path using Möbius addition:
\begin{equation}
\mathbf{h}_i^{\text{HCOE}}
= \bigoplus_{\ell=1}^{L} \mathbf{h}_{a_i^{(\ell)}}^{(\ell)}
\end{equation}
where $\bigoplus$ denotes sequential Möbius addition from the coarsest
to the finest ontology level ($\ell=1$ to $L$), keeping the aggregated
representation inside the Poincaré ball. The resulting representation $\mathbf{h}_i^{\text{HCOE}}$ integrates biomedical LM semantics with the coarse-to-fine structure of the clinical ontology.


\section{Experiments}


\subsection{MIMIC-IV Dataset and Preprocessing}

We evaluated HCOE on MIMIC‑IV, a publicly available, de-identified EHR database from the Beth Israel Deaconess Medical Center \cite{johnson2023mimic}. The database contains comprehensive longitudinal hospital and intensive care data collected from emergency department and inpatient admissions between 2008 and 2022. We use the inpatient hospitalization data, including diagnoses, procedures, medication prescriptions, and admission and discharge timestamps. We extracted clinical codes from all patients and represented them using their corresponding five-level codes. For each patient, all clinical codes within a visit are treated as an unordered set, while visits are chronologically ordered. After preprocessing, the cohort contains 223,452 patients and 546,028 hospital admissions, with 9,143 ICD-9 diagnosis codes, 19,440 ICD-10 diagnosis codes, 2,557 ICD-9 procedure codes, 12,354 ICD-10 procedure codes, and 1,286 ATC level-5 medication codes.

\subsection{ICD/ATC Relation Prediction and PheCode Transfer}

To evaluate whether HCOE captures clinical relationships in biomedical taxonomies, we assess its performance across four clinical relation prediction settings: ICD$\rightarrow$ICD, ATC$\rightarrow$ATC, ICD$\rightarrow$ATC, and CCS$\rightarrow$PheCode. We first perform multi-hop relation prediction within the ICD and ATC hierarchies to predict ancestor–descendant relations beyond directly observed parent-child links. For each positive relation pair, we sample ten negative candidates, using sibling concepts as hard negatives when available and otherwise sampling unrelated concepts uniformly at random. We also evaluate a cross-ontology one-hop relation prediction task between medications and diagnoses using ICD-10$\rightarrow$ATC links~\cite{Zou2022-te}. Given an ATC medication code, the task is to  predict whether a candidate ICD-10 diagnosis code is associated with it. Since each ATC code is paired with one ICD-10 label in our dataset, we sample one negative ICD-10 code for each positive pair. Finally, we evaluate HCOE on an unseen PheCode-based ICD hierarchy after training on the CCS-based ICD ontology~\cite{Yang2026-ls}. For each query ICD code, candidates sharing its integer-level PheCode are positives, whereas candidates from other PheCodes are hard negatives. We compare candidates by hyperbolic distance and test whether positives rank above negatives. 
For each method, the decision threshold is selected on the validation set and fixed for the test set.
All reported Precision, Recall, and F1-score values are computed on the held-out test set.
We compare HCOE with four groups of baselines: (1) a hierarchy-agnostic LM baseline BioBERT~\cite{lee2020biobert}; (2) structure-aware LM baselines, including SapBERT~\cite{liu2021learning}, HiTs \cite{he2024language}, and OnT~\cite{OnT}; (3) structure-aware embedding baselines, including RotatE~\cite{sun2019rotate} and cui2vec~\cite{beam2020clinical}; and (4) hyperbolic embedding baselines, including Poincaré Embedding \cite{nickel2017poincare}, Hyperbolic Entailment Cone \cite{pmlr-v80-ganea18a}, and Poincaré GloVe embedding \cite{tifrea2018poincare}. 

\subsection{MIMIC-IV Downstream Prediction Setup}

For clinical prediction, we extracted diagnosis, procedure, and medication codes from MIMIC-IV and evaluated three common tasks, including mortality and readmission prediction as well as drug recommendation. 
Mortality prediction assesses whether a patient passes away within 90 days after discharge. Readmission prediction assesses whether a patient is readmitted within the next 15 days after discharge. Drug recommendation predicts the set of medications for each visit given the patient’s prior clinical history~\cite{wang2026saferxagent}. Mortality and readmission are binary classification tasks evaluated with Area Under Precision-Recall Curve (AUPRC) and Area Under Receiver Operating Characteristic Curve (AUROC). Drug recommendation is a multi-label prediction task evaluated with AUPRC, F1-score, and Jaccard. We also predicted rare drugs on 98 ATC codes with frequencies between 10 and 20. For each of the 98 rare drugs, we compute Recall@15 over visits containing that drug. For all downstream tasks, we split patients into training, validation, and test sets using a 70\%/10\%/20\% split.

To facilitate a controlled comparison of embedding quality, all methods are evaluated using the same mean-pooling strategy and linear prediction head.
We map clinical codes to their HCOE embeddings and use mean pooling over code embeddings to obtain a patient-level representation. For all downstream tasks, we use a single linear prediction head on top of the patient-level representation. Mortality and readmission are modeled as binary classification tasks with a sigmoid activation and binary cross-entropy loss. Drug recommendation is modeled as a multi-label classification task, where a linear layer outputs logits for all medication codes and an element-wise sigmoid activation is applied. The model is trained using binary cross-entropy loss over all medication labels. The best model is selected according to the validation loss on the target task.

We compare HCOE with three groups of baselines: (1) structure-aware embedding baselines, including GRAM \cite{choi2017gram}, cui2vec \cite{beam2020clinical}, and RotatE \cite{sun2019rotate}; (2) biomedical LM baselines, including BioBERT \cite{lee2020biobert}, G-BERT \cite{shang2019pre}, ClinicalBERT \cite{alsentzer2019publicly}, and BEHRT \cite{li2020behrt}; (3) structure-aware LM baselines, including SapBERT~\cite{liu2021learning}, HiTs~\cite{he2024language}, and OnT~\cite{OnT}.

\begin{table*}[t]
\centering
\caption{Clinical prediction results on MIMIC-IV for mortality, readmission, drug recommendation, and rare drug prediction tasks. Metrics are reported as mean (standard error) from bootstrap. The best results in each column are highlighted in bold.}
\label{tab:tasks_mimic4}
\begin{tabular}{lcccccccc}
\toprule
\textbf{Method} &
\multicolumn{2}{c}{\textbf{Mortality}} &
\multicolumn{2}{c}{\textbf{Readmission}} &
\multicolumn{3}{c}{\textbf{Drug}} &
\multicolumn{1}{c}{\textbf{Rare Drug}} \\
\cmidrule(lr){2-3}\cmidrule(lr){4-5}\cmidrule(lr){6-8}\cmidrule(lr){9-9}
& \textbf{AUPRC} & \textbf{AUROC}
& \textbf{AUPRC} & \textbf{AUROC}
& \textbf{AUPRC} & \textbf{F1} & \textbf{Jaccard}
& \textbf{Recall@15} \\
\midrule
\textbf{HCOE} & \textbf{44.3 (0.8)} & \textbf{87.2 (1.1)} & \textbf{48.7 (0.6)} & \textbf{72.3 (0.6)} & \textbf{76.3 (0.1)} & \textbf{64.3 (0.2)} & \textbf{50.7 (0.2)} & \textbf{44.2 (0.3)} \\
\midrule
GRAM & 38.5 (0.7) & 79.1 (0.9) & 43.2 (0.8) & 68.2 (0.4) & 72.0 (0.3) & 61.3 (0.4) & 48.6 (0.4) & 34.5 (0.2) \\
cui2vec  & 39.0 (0.4) & 80.4 (0.8) & 43.7 (0.6) & 68.9 (0.3) & 75.2 (0.1) & \textbf{64.3 (0.3)} & 49.9  (0.4) & 36.6 (0.1) \\
RotatE  & 39.2 (0.3) & 79.6 (0.6) & 43.7 (0.4) & 69.2 (0.6) & 74.1 (0.2) & 62.7 (0.4) & 48.7 (0.4) & 37.4 (0.1) \\
BioBERT & 43.2 (0.9) & 86.1 (1.3) & 46.5 (0.9) & 71.5 (1.2) & 73.9 (0.1)& 60.3 (0.2) & 44.5 (0.2) & 40.6 (0.2) \\
G-BERT & 42.7 (0.4) & 85.0 (1.3) & 47.3 (0.4) & 70.1 (0.7) & 75.4(0.1)& 61.4 (0.3) & 45.5 (0.3) & 42.7 (0.3)  \\
BEHRT  & 42.0 (0.3) & 84.3 (0.7) & 44.5 (0.4) &  69.0 (0.8) & 74.3 (0.2) & 60.2 (0.2) & 45.7 (0.3) &37.5 (0.3)  \\
ClinicalBERT & 43.5 (0.9) & 86.2 (0.8) & 45.7 (0.6) & 70.4 (0.6) & 73.8 (0.2) &59.4(0.4) & 44.0 (0.4) & 38.4 (0.4) \\
SapBERT  & 42.4 (0.6) & 83.7 (1.2) & 45.3 (0.7) & 70.2 (1.1)  & 74.6 (0.1) & 60.5 (0.2) & 45.7(0.4) & 42.0 (0.3) \\
HiTs & 42.7 (0.6) & 81.5 (0.7) & 45.2 (0.4) & 69.6 (0.8) & 74.8 (0.2)& 61.6 (0.2) & 48.0 (0.1) & 37.5 (0.1) \\
OnT & 43.3 (0.5) & 83.4 (1.1) & 45.9 (0.7) &70.3 (0.8)  & 71.0 (0.1) & 57.3 (0.2) & 44.5 (0.1) &35.8 (0.1) \\
\bottomrule
\end{tabular}
\end{table*}

\subsection{Implementation Details}

We use BioBERT as the frozen biomedical LM backbone and obtain concept embeddings from its pooled output
 \cite{lee2020biobert}. BioBERT produces $d_0=768$-dimensional embeddings, which are projected to $d_h=256$-dimensional hyperbolic embeddings. We set the curvature parameter of the Poincar\'e ball to 
$\kappa = 1/d_h$ and the margin parameter in the hyperbolic triplet loss to $\alpha=5$. All models were implemented in PyTorch and optimized with Adam optimizer with a learning rate of $1\times10^{-4}$ and weight decay of $1\times10^{-5}$. The training process uses a batch size of 64 for up to 20 epochs with early stopping.

\begin{figure}[t]
    \centering
    \includegraphics[width=\columnwidth]{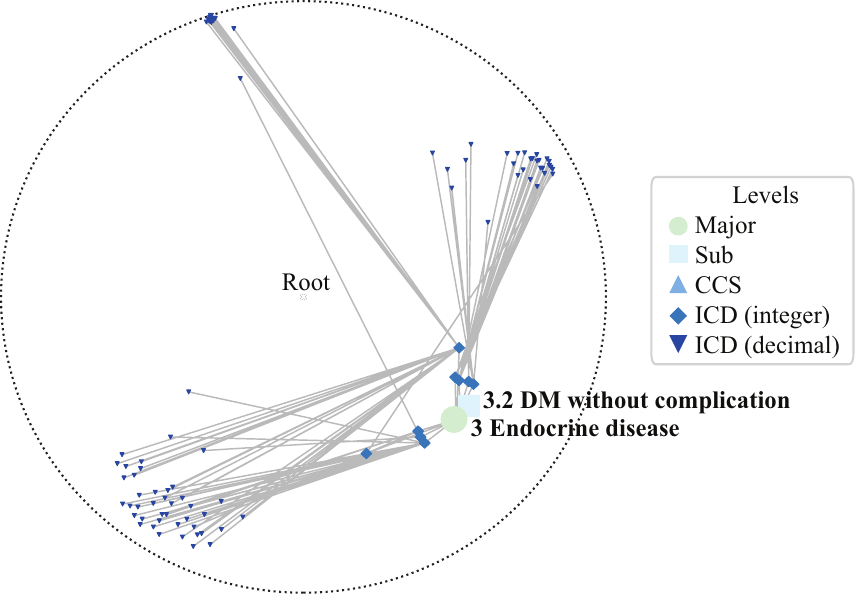}    \caption{Visualization of HCOE embeddings for ICD diagnosis codes under the endocrine disease category (major category 3) using a Poincaré map. The embeddings of the endocrine disease category exhibit a radial hierarchy, with coarse concepts near the center and fine-grained codes toward the boundary.
    } 
    \label{fig:emb_vis}
    \vspace{-1.5\baselineskip} 
\end{figure}

\section{Results}

\subsection{HCOE Preserves Clinical Ontology Relations}

As shown in Table~\ref{tab:hier_pred_prf}, HCOE achieves the best F-score across all four clinical relation prediction tasks. For within-ontology prediction, HCOE obtains F-scores of 75.6\% on ICD$\rightarrow$ICD and 76.3\% on ATC$\rightarrow$ATC, corresponding to absolute gains of 8.5\% and 1.8\% over the strongest baselines, respectively. These results indicate that combining biomedical LM semantics with hyperbolic ontology contrastive learning better preserves ancestor-descendant relations in expert-defined clinical taxonomies. HCOE also outperforms BioBERT, indicating that hyperbolic contrastive learning better distinguishes true ontology relations from unrelated candidate pairs than textual similarity alone. HCOE further outperforms structure-aware embedding methods, showing that combining biomedical LM semantics with clinical ontology yields more discriminative concept representations. It also surpasses hyperbolic embedding baselines, indicating that hyperbolic geometry alone is insufficient without LM-based semantics. Cross-ontology prediction is more challenging because diagnosis and medication concepts come from separate taxonomies, and within-ontology training does not directly provide alignment between ICD and ATC concepts. Nevertheless, HCOE still ranks first with F-score 56.7\%, outperforming the strongest baseline HiTs with F-score 53.2\%. These results suggest that HCOE can support relation prediction across heterogeneous clinical coding systems, where diagnosis and medication concepts come from different taxonomies. On the CCS$\rightarrow$PheCode transfer task, HCOE also achieves the highest F-score  80.6\%, outperforming the strongest baseline HiTs (76.5\%) by 4.1\%. Because PheCode labels are not used during training, this result suggests that HCOE learns clinical relationships among ICD diagnostic codes that generalize beyond the CCS-based training hierarchy to an unseen PheCode-based hierarchy.

\begin{table*}[t]
\centering
\setlength{\tabcolsep}{4.5pt}
\caption{Ablation of HCOE on four clinical relation prediction tasks, including ICD$\rightarrow$ICD, ATC$\rightarrow$ATC, ICD$\rightarrow$ATC, and CCS$\rightarrow$PheCode. We report F-score for each task. The best results in each column are highlighted in bold.}
\label{tab:ablation}
\begin{tabular}{lcccccc}
\toprule
\textbf{Variant} 
& \textbf{Path aggregation} 
& \textbf{Contrastive objective}
& \textbf{ICD$\rightarrow$ICD}
& \textbf{ATC$\rightarrow$ATC}
& \textbf{ICD$\rightarrow$ATC}
& \textbf{CCS$\rightarrow$PheCode}\\
\midrule
Full HCOE 
& Yes 
& Parent + Child 
& 75.6 
& 76.3 
& 56.7
& 80.6 \\
w/o path aggregation 
& No 
& Parent + Child 
& 74.3 
& 75.0 
& 55.1
& 78.6 \\
Parent-side only 
& Yes 
& Parent only 
& 71.7 
& 72.7 
& 55.8 
& 75.0 \\
Child-side only 
& Yes 
& Child only 
& 63.5 
& 67.4 
& 52.5
& 73.5 \\
w/o contrastive learning 
& Yes 
& None 
& 47.6 
& 51.6 
& 51.2
& 71.5 \\
\bottomrule
\end{tabular}
\end{table*}

\subsection{Qualitative Visualization of HCOE Representations}

To examine whether HCOE learns hierarchy-consistent geometry, we visualize the embeddings of all descendant concepts under the endocrine disease category (major category 3). We use a Poincaré map, a two-dimensional visualization of hyperbolic embeddings, to illustrate all ontology nodes under the endocrine disease category \cite{poincare_map}. As shown in Fig.~\ref{fig:emb_vis}, the visualization shows a clear radial structure: coarse-grained concepts, such as 3 endocrine disease and 3.2 diabetes mellitus without complication, lie near the center of the Poincaré ball, whereas fine-grained ICD codes  are positioned closer to the boundary.
This pattern is consistent with hyperbolic geometry, where the available volume grows exponentially with radius and provides greater capacity near the boundary for fine-grained concepts. This result suggests that HCOE captures the coarse-to-fine structure of the clinical ontology while producing interpretable hyperbolic concept embeddings.

\subsection{HCOE Improves Clinical Prediction}
\label{sec: pred_results}

As shown in Table~\ref{tab:tasks_mimic4}, HCOE consistently improves clinical prediction performance across all downstream clinical prediction metrics on MIMIC-IV. For mortality and readmission, HCOE outperforms the strongest baseline in mortality (AUPRC 44.3\% vs. 43.5\% and AUROC 87.2\% vs. 86.2\%) and in readmission (AUPRC 48.7\% vs. 47.3\% and AUROC 72.3\% vs. 71.5\%). These results indicate that hierarchy-aware clinical concept embeddings produce more discriminative patient-level features than pre-trained biomedical LM representations alone. The gains are most evident on medication recommendation tasks. For drug recommendation, HCOE achieves the best scores across all three metrics, indicating that incorporating clinical ontology structure helps capture relationships among medication codes, leading to more effective multi-label drug prediction. For rare drug prediction, HCOE achieves the highest Recall@15 (44.2\%), outperforming the best baseline (42.7\%). Because rare drugs have limited training examples, this improvement suggests that HCOE can leverage shared ontology structure among related medications to improve retrieval of long-tail drug codes.

\subsection{Ablation Studies}

We conduct ablation studies to isolate the effects of path aggregation and hyperbolic contrastive learning in HCOE. 
As shown in Table~\ref{tab:ablation}, we first evaluate the role of path aggregation by removing it while keeping the full hyperbolic contrastive learning objective. We then evaluate the role of the contrastive objective by keeping path aggregation and using only the parent-side objective, only the child-side objective, or no contrastive learning. We report F-score on four clinical relation prediction tasks: ICD$\rightarrow$ICD, ATC$\rightarrow$ATC, ICD$\rightarrow$ATC, and CCS$\rightarrow$PheCode.

As shown in Table~\ref{tab:ablation}, the full HCOE model achieves the best overall performance on all four tasks, demonstrating the effectiveness of jointly using path aggregation and hyperbolic contrastive learning for clinical concept representation. Removing the hyperbolic contrastive learning objective causes the largest performance drop, reducing the F-score from 75.6 to 47.6 on ICD$\rightarrow$ICD and from 76.3 to 51.6 on ATC$\rightarrow$ATC. 
This indicates that removing the hyperbolic contrastive objective substantially weakens the hierarchy-aware representations, whereas using both parent-side and child-side contrastive objectives produces more hierarchy-consistent hyperbolic embeddings. Comparing the two one-sided contrastive objectives, the parent-side objective consistently outperforms the child-side objective across all tasks, suggesting that parent concepts serve as more stable anchors for organizing fine-grained clinical codes. Moreover, the full contrastive objective consistently outperforms both one-sided variants, indicating that parent-side and child-side supervision provide complementary hierarchical information. Overall, these results demonstrate that HCOE improves clinical relation prediction by integrating biomedical LM semantics, path-level ontology aggregation, and the proposed hyperbolic contrastive learning.

\subsection{Sensitivity Analysis of  Dimension }

\begin{table}[t]
\centering
\setlength{\tabcolsep}{4.5pt}
\caption{
Sensitivity of HCOE to the hyperbolic latent dimension. 
We report F-score on clinical relation prediction tasks.
}
\label{tab:dim_sensitivity}
\begin{tabular}{lccc}
\toprule
\textbf{$d_h$} 
& \textbf{ICD$\rightarrow$ICD} 
& \textbf{ICD$\rightarrow$ATC} 
& \textbf{CCS$\rightarrow$PheCode} 
 \\
\midrule
128 & 74.4 & 56.1 & 79.2  \\
256 & 75.6 & 56.7 & 80.6  \\
768 & 72.7 & 56.3 & 77.4 \\
\bottomrule
\end{tabular}
\end{table}

We assess the sensitivity of HCOE to the hyperbolic embedding dimension using three settings, $d_h \in \{128,256,768\}$, on three representative clinical relation prediction tasks: ICD$\rightarrow$ICD, ICD$\rightarrow$ATC, and CCS$\rightarrow$PheCode. As shown in Table~\ref{tab:dim_sensitivity}, HCOE achieves the best performance with $d_h=256$ on all three tasks. The original LM embedding size, $d_h=768$, does not provide additional gains and may introduce unnecessary complexity, whereas $d_h=128$ appears to limit representation capacity. The ICD$\rightarrow$ATC results vary only slightly across dimensions, suggesting that cross-ontology prediction is less sensitive to embedding dimension. We thus use $d_h=256$ for all experiments, as it achieves the best overall performance.

\section{Conclusion}


HCOE combines frozen BioBERT representations, ontology-guided hyperbolic contrastive learning, and coarse-to-fine path aggregation for hierarchy-aware medical code embeddings. Evaluations cover ICD/ATC clinical relation prediction, CCS-to-PheCode hierarchy transfer, and MIMIC-IV mortality prediction, readmission prediction, medication recommendation, and rare drug prediction. The reported results support its use for clinical ontology representation and downstream EHR modeling, with ablations assessing contrastive learning and path aggregation. This study focuses on common clinical ontologies and standard EHR prediction tasks, leaving broader biomedical knowledge sources for future exploration. Future work will explore richer biomedical knowledge graphs, including UMLS, and multimodal clinical inputs such as clinical notes.




\bibliographystyle{IEEEtran}
\bibliography{references}

\end{document}